\documentclass[sigconf, nonacm]{acmart}

\usepackage{algorithm}
\usepackage{algpseudocode}
\usepackage{multirow}
\usepackage{graphicx}
\AtBeginDocument{%
  }

\setcopyright{acmlicensed}
\copyrightyear{2026}
\acmYear{2026}
\acmDOI{XXXXXXX.XXXXXXX}
\acmConference[GECCO '26]{Genetic and Evolutionary Computation Conference}{July 14--18, 2026}{San Jos\'{e}, Costa Rica}
\acmISBN{XXX-X-XXXX-XXXX-X/2026/07}

\begin{document}

\title{Associative Constructive Evolution: Enhancing Metaheuristics through Hebbian-Learned Generative Guidance}

\author{Shanxian Lin}
\affiliation{%
  \institution{Graduate School of Technology, Industrial and Social Sciences, Tokushima University}
  \city{Tokushima}
  \postcode{770-8506}
  \country{Japan}}
\email{utigooselsx@gmail.com}

\author{Yuichi Nagata}
\affiliation{%
  \institution{Graduate School of Technology, Industrial and Social Sciences, Tokushima University}
  \city{Tokushima}
  \postcode{770-8506}
  \country{Japan}}
\email{nagata@is.tokushima-u.ac.jp}

\author{Haichuan Yang}
\affiliation{%
  \institution{Graduate School of Technology, Industrial and Social Sciences, Tokushima University}
  \city{Tokushima}
  \postcode{770-8506}
  \country{Japan}}
\email{you.kaisen@tokushima-u.ac.jp}

\begin{abstract}
Metaheuristic algorithms such as Particle Swarm Optimization (PSO) and Evolutionary Algorithms (EA) excel at exploring solution spaces but lack mechanisms to accumulate and reuse procedural knowledge from successful search trajectories. This paper proposes Associative Constructive Evolution (ACE), a framework that enhances metaheuristics through learned generative guidance. ACE introduces a Generative Construction Automaton (GCA)---a probabilistic model over operation sequences---coupled with the base metaheuristic in a synergistic loop: the metaheuristic explores and provides trajectory samples, while the GCA consolidates successful patterns and guides future exploration. Three mechanisms realize this cooperation: Hebbian weight consolidation that strengthens associations between co-successful operations, guided sampling that biases search toward learned high-quality regions, and symbolic abstraction that extracts frequent patterns into reusable macro-operations. Experiments integrating ACE with EA and PSO on molecular design and maze navigation demonstrate consistent improvements. ACE-PSO achieves a 27.5\% increase in success rate while reducing convergence time by 49.6\%. In molecular design, ACE-EA improves fitness by 10.1\% with 126 chemically interpretable macro-operations automatically discovered.
\end{abstract}




\maketitle

\section{Introduction}

Metaheuristic algorithms---including Evolutionary Algorithms (EA) and Particle Swarm Optimization (PSO)---excel at solution space exploration through population-based mechanisms~\cite{eiben2015introduction, 488968}. However, they suffer from a fundamental ``exploration without memory'' limitation: procedural knowledge accumulated during search is typically discarded upon termination. For instance, in pharmaceutical optimization, a swarm might repeatedly discover that specific modification sequences improve stability, yet this structural insight remains implicit in transient particle positions and is lost when the run concludes~\cite{Polishchuk2020}. This reflects a missing architectural component: while these algorithms excel as \textit{explorers}, they lack a complementary \textit{cartographer} to map successful search trajectories and guide future exploration.

We propose Associative Constructive Evolution (ACE), a framework that provides this missing cartographic capability. ACE introduces a Generative Construction Automaton (GCA)---a probabilistic generative model that learns to produce successful operation sequences through collective exploration experience. Unlike classical Evolutionary Automata where each individual is a distinct state machine, ACE employs a collective learning paradigm: the entire population trains a single, shared automaton that captures the generative ``grammar'' of successful construction.

The explorer-cartographer cooperation is realized through three mechanisms. \textbf{Hebbian weight consolidation} updates GCA weights when operations co-occur in successful trajectories, following the principle that ``neurons that fire together wire together''~\cite{Hebb2002}. \textbf{Guided sampling} biases the metaheuristic's search toward learned high-quality regions while maintaining exploration through temperature-controlled stochasticity. \textbf{Symbolic abstraction} extracts operation pairs that persistently co-occur in successful trajectories into macro-operations---composite primitives that can be invoked as atomic units.

ACE shares conceptual roots with Estimation of Distribution Algorithms (EDAs) and Ant Colony Optimization (ACO) but maintains a distinct architectural role. While EDAs typically replace genetic operators with probabilistic sampling~\cite{larranaga2001estimation, 4129846}, ACE retains the base metaheuristic's variation mechanisms, using the learned model strictly for guidance. Unlike ACO's scalar pheromones, ACE employs symbolic abstraction to crystallize patterns into discrete macro-operations, dynamically altering the search space topology---a capability absent in standard probabilistic models. ACE also connects to knowledge transfer in evolutionary optimization~\cite{9321762} and hierarchical skill discovery in reinforcement learning~\cite{SUTTON1999181}.

The contributions of this paper are threefold. First, we formalize the explorer-cartographer framework, establishing how generative guidance models can enhance metaheuristic search. Second, we introduce Hebbian weight consolidation as a gradient-free mechanism for learning operation associations from exploration trajectories. Third, we demonstrate through experiments that ACE consistently enhances EA and PSO, with improvements reaching 27.5\% in success rate for ACE-PSO ($p < 0.0001$), and validate that long-term evolution enables macro-operation solidification with chemically interpretable patterns.

\section{Methodology}

\subsection{Explorer-Cartographer Architecture}

Let $\mathcal{O} = \{o_1, o_2, \ldots, o_n\}$ denote the set of atomic operations, and let $\Pi$ denote the space of operation sequences. An execution function $\text{exec}: \Pi \times \mathcal{S}_0 \rightarrow \mathcal{S}$ maps an initial state $s_0$ and program $\pi = (o_{i_1}, \ldots, o_{i_k})$ to a solution. The explorer generates trajectory samples and provides fitness evaluations; different metaheuristics generate trajectories through different mechanisms (EA through crossover/mutation, PSO through velocity-based updates).

The cartographer is realized as a \textbf{Generative Construction Automaton (GCA)}. For a trajectory $\pi = (o_{i_1}, \ldots, o_{i_k})$, the GCA defines transition probabilities parameterized by a learned weight matrix $\mathbf{W} \in \mathbb{R}^{n \times n}$:
\begin{equation}
P_\Theta(o_j \mid o_i) = \frac{\exp(W_{ij} / \tau)}{\sum_{l \in \mathcal{N}(i)} \exp(W_{il} / \tau)}
\label{eq:transition_prob}
\end{equation}
where $\mathcal{N}(i)$ denotes valid successor operations for $o_i$, and temperature $\tau > 0$ controls exploration-exploitation balance. The GCA additionally maintains a support matrix $\mathbf{S} \in \mathbb{N}^{n \times n}$ recording co-occurrence frequencies for abstraction decisions, and a macro-operation library $\mathcal{M}$. High weights indicate transitions that have historically appeared in successful trajectories; the softmax converts these into valid probability distributions. For large operation spaces, sparse storage reduces memory from $O(n^2)$ to $O(|\mathcal{T}|)$ where $\mathcal{T}$ is the valid transition set.

\begin{figure}[htbp]
 \centering
 \includegraphics[width=\columnwidth]{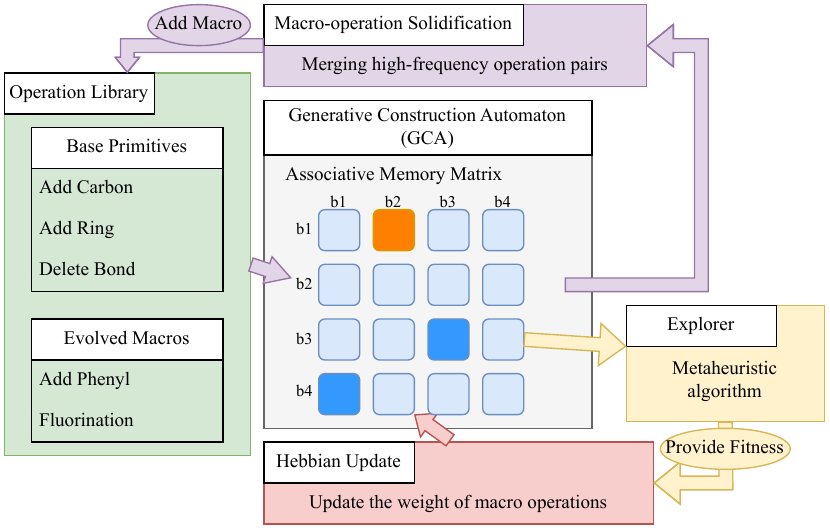}
 \caption{Overview of the ACE framework. Purple arrows denote the main optimization loop, while the remaining arrows signify the knowledge abstraction pathway.}
 \label{fig:ace_framework}
\end{figure}

The cooperation loop operates as: (1) the metaheuristic generates trajectories, optionally biased by GCA; (2) trajectories are evaluated; (3) successful trajectories update GCA via Hebbian consolidation; (4) the GCA generates biases for subsequent exploration. Figure~\ref{fig:ace_framework} illustrates this architecture.

\subsection{Hebbian Weight Consolidation}

Hebbian consolidation is the core learning mechanism. When the metaheuristic produces offspring from parent trajectories, let $\mathbf{x}_A, \mathbf{x}_B \in \mathbb{R}^n$ denote operation count vectors. The fitness gain is $\Delta f = f_{\text{child}} - \frac{1}{2}(f_A + f_B)$. The weight matrix updates:
\begin{equation}
\mathbf{W}^{(t+1)} = (1 - \gamma)\mathbf{W}^{(t)} + \lambda \cdot [\Delta f]_+ \cdot (\mathbf{x}_A \mathbf{x}_B^\top + \mathbf{x}_B \mathbf{x}_A^\top)
\label{eq:hebbian_update}
\end{equation}
where $\gamma \in [0, 1]$ is the decay factor, $\lambda > 0$ the learning rate, and $[\cdot]_+ = \max(\cdot, 0)$ ensures only positive fitness gains trigger learning. This gradient-free rule reinforces associations between co-successful operations---applicable to black-box fitness functions without requiring analytical derivatives. It ensures \emph{locality} (only active weights update) and \emph{fitness modulation} (learning scales with improvement).

\subsection{Guided Sampling}

The GCA biases operation selection via learned transition probabilities (Eq.~\ref{eq:transition_prob}). To prevent premature convergence, a minimum probability floor ensures continued exploration:
\begin{equation}
P'_\Theta(o_j \mid o_i) = (1 - \epsilon) \cdot P_\Theta(o_j \mid o_i) + \frac{\epsilon}{|\mathcal{N}(i)|}
\label{eq:exploration_floor}
\end{equation}
where $\epsilon \in (0, 1)$ is the exploration floor parameter. Temperature $\tau$ provides additional control: high temperature produces near-uniform generation; low temperature concentrates on high-weight transitions. This guidance respects the base metaheuristic's dynamics---for PSO, biases influence velocity updates; for EA, guidance affects mutation selection.

\subsection{Symbolic Abstraction}

When operation pairs persistently co-occur in successful trajectories, ACE abstracts them into macro-operations. Abstraction triggers when three conditions hold for pair $(o_i, o_j)$:
(1) association strength $W_{ij} > \theta_W$;
(2) statistical support $S_{ij} \geq \theta_S$;
(3) specificity $\text{lift}_{ij} = W_{ij} / (\bar{W}_{\cdot i} \cdot \bar{W}_{j \cdot}) \geq \theta_L$.
The lift condition measures whether the observed association significantly exceeds the product of marginal average weights (the baseline expectation under independence), avoiding abstraction of universally common patterns. Upon creation, a macro-operation joins the vocabulary as a new atomic primitive. The weight matrix expands to $\mathbf{W}' \in \mathbb{R}^{(n+1) \times (n+1)}$ with transition weights initialized from constituent averages. This introduces variable-length encoding: a single decision step now executes a multi-step sequence, enabling the search to traverse the landscape in larger, semantically meaningful strides. Periodic pruning removes macro-operations whose usage-weighted success rate falls below threshold $\theta_{\text{eff}}$, preventing unbounded vocabulary growth.

\subsection{Integration with Metaheuristics}

Algorithm~\ref{alg:ace} presents the ACE-enhanced optimization loop.

\begin{algorithm}[htbp]
\caption{ACE-Enhanced Metaheuristic Optimization}
\label{alg:ace}
\begin{algorithmic}[1]
\Require Base metaheuristic $\mathcal{A}$, population size $N$, max generations $G$
\State Init population $\mathcal{P}$ via $\mathcal{A}$; init GCA: $\mathbf{W} \leftarrow \mathbf{0}$, $\mathcal{M} \leftarrow \emptyset$
\For{$t = 1$ to $G$}
    \State Evaluate fitness for each trajectory in $\mathcal{P}$
    \State $\mathcal{A}$ generates offspring via its operators
    \For{each successful recombination ($\Delta f > 0$)}
        \State Update $\mathbf{W}$ via Eq.~(\ref{eq:hebbian_update})
    \EndFor
    \State Bias new trajectories using Eq.~(\ref{eq:transition_prob})
    \If{$t \bmod T_{\text{abstract}} = 0$}
        \State Form qualifying macro-operations; prune ineffective ones
    \EndIf
    \State Update population via $\mathcal{A}$'s selection mechanism
\EndFor
\State \Return Best solution, trained GCA $\{\mathbf{W}^*, \mathcal{M}^*\}$
\end{algorithmic}
\end{algorithm}

The modular interface requires only that the metaheuristic provides trajectories and fitness; ACE returns sampling biases. Each metaheuristic retains its characteristic exploration dynamics while gaining cumulative learning capability.

\section{Experiments}

We validate ACE by integrating it with EA and PSO across molecular design and maze navigation. All experiments were conducted on AMD Ryzen 7 9700X 8-Core CPU. Thresholds ($\theta_W{=}0.3$, $\theta_S{=}3$, $\theta_L{=}1.4$) were fixed across all experiments without instance-specific tuning. Sensitivity analysis is planned for future work.

\subsection{Molecular Design}

\subsubsection{Setup and Results.}
Experiments extend the MCEMOL framework~\cite{Lin2026} with 37 chemical operations from CReM~\cite{Polishchuk2020}. The fitness function integrates 11 components following the original weighting scheme~\cite{Lin2026}. Configuration: 100 generations, population 100, $\lambda{=}0.15$, $\gamma{=}0.2$, crossover 0.3, mutation 0.4, elitism 10\%. All parameters were fixed across runs without tuning.

\begin{table}[htbp]
  \caption{Molecular Design: Long-term Evolution Results (5 Runs, 100 Generations)}
  \label{tab:mcemol_results}
  \begin{tabular}{lcc}
    \toprule
    Metric & ACE-EA & Standard EA \\
    \midrule
    Final Fitness & 3.9010 & 3.5438 \\
    Fitness Improvement & \multicolumn{2}{c}{+10.1\%} \\
    \midrule
    Valid Molecules Generated & 1,641 & 834 \\
    Molecule Generation Improvement & \multicolumn{2}{c}{+96.8\%} \\
    \midrule
    Convergence to Std Final (Gen) & 49 & 100 \\
    Convergence Speedup & \multicolumn{2}{c}{2.04$\times$} \\
    \midrule
    Macro-operations Solidified & 126 & --- \\
    Runtime (hours) & 62.95 & 35.41 (OH: 77.8\%) \\
    \bottomrule
  \end{tabular}
\end{table}

ACE-EA achieves 10.1\% higher final fitness (3.9010 vs.\ 3.5438) while generating 96.8\% more valid molecules (Table~\ref{tab:mcemol_results}). Notably, ACE-EA reaches the standard EA's final fitness level at generation 49, representing a 2.04$\times$ convergence speedup. While the runtime overhead appears substantial (77.8\%), this reflects the use of a fast surrogate fitness function; in practical scenarios with expensive evaluations (e.g., DFT simulations), GCA maintenance overhead becomes negligible compared to evaluation cost.

\subsubsection{Macro-operation Analysis.}
The 100-generation horizon enables substantial macro-operation solidification: 126 macro-operations emerged from 918 Hebbian updates. Ring modifications dominate (43, 34.1\%), followed by halogenation (36, 28.6\%) and heterocycle additions (29, 23.0\%). These encode interpretable medicinal chemistry principles: e.g., MACRO10 (\texttt{add\_boronic\_acid + add\_piperazine}) combines a Suzuki coupling handle with a solubility-enhancing group; MACRO0 (\texttt{add\_fluorine + replace\_benzene\_with\_pyridine}) combines metabolic blocking with bioisosteric modulation.

\subsubsection{Transfer Validation.}
Injecting the 126 discovered patterns into a warm-started run reached the cold-start optimum ($f{=}3.90$) by generation 57 (vs.\ 100) and achieved a higher maximum of $f{=}4.23$, confirming that macro-operations encode reusable procedural knowledge rather than spurious correlations. The chemical interpretability of these patterns---aligning with established dual-objective optimization strategies in drug design---further validates that ACE's symbolic abstraction extracts genuine domain structure.

\subsection{Maze Navigation}

\subsubsection{Setup.}
We use the Maze Benchmark for Evolutionary Algorithms~\cite{10.1145-3205651.3208285} with 40 SCMP mazes ($15{\times}15$) across four connectivity levels: 0\%, 30\%, 60\%, 100\%. Four configurations were evaluated: Standard EA/PSO vs.\ ACE-EA/PSO.

For discrete PSO, velocity is redefined as a selection probability over valid neighbors. For a particle at node $u$, the probability of moving to neighbor $v$ integrates PSO memories with ACE guidance:
\begin{equation}
S(v) = w \delta(v, \pi_{prev}) + c_1 r_1 \delta(v, \pi_{pbest}) + c_2 r_2 \delta(v, \pi_{gbest}) + \alpha h(v) + \lambda P_{\Theta}(v \mid u)
\end{equation}
where $\delta$ indicates alignment with previous, personal best, or global best paths, $h(v)$ provides goal-directed heuristic bias applied identically to both standard and ACE configurations, and $P_\Theta$ is the learned GCA probability. Since PSO lacks crossover, the GCA update uses a simplified single-trajectory formulation.

Configuration: 30 runs per algorithm-maze combination (1200 total), population 100, max 500 generations, uniform parameters across all 40 mazes.

\subsubsection{Overall Results.}

\begin{table}[htbp]
\centering
\caption{ACE Enhancement: Overall Maze Results (1200 Runs per Config)}
\label{tab:maze_overall}
\resizebox{\columnwidth}{!}{%
    \begin{tabular}{lccccc}
    \toprule
    Configuration & Succ.\ Rate & Mean Fit. & Mean Gen. & Path Eff. & Time (s) \\
    \midrule
    Standard EA & 62.0\% & 9541.9 & 233.9 & 0.900 & 0.55 \\
    ACE-EA & 65.8\% & 9578.8 & 212.7 & 0.904 & 0.84 \\
    \midrule
    Standard PSO & 55.1\% & 9496.5 & 256.2 & 0.896 & 10.97 \\
    ACE-PSO & \textbf{82.6\%} & \textbf{9711.7} & \textbf{129.0} & \textbf{0.916} & 20.55 \\
    \bottomrule
    \end{tabular}%
}
\end{table}

ACE consistently enhances both metaheuristics (Table~\ref{tab:maze_overall}). The most dramatic improvement occurs with PSO: ACE-PSO achieves 82.6\% success rate vs.\ 55.1\% for standard PSO (+27.5 points), with convergence accelerating by 49.6\% (129.0 vs.\ 256.2 generations). This demonstrates strong synergy between PSO's particle-based exploration and ACE's learned guidance. ACE-EA shows more modest gains (+3.8 points), as the baseline EA already possesses reasonable exploitation through selection pressure. This asymmetry reveals a design principle: ACE provides greatest benefit to metaheuristics with strong exploration but weak exploitation, where the cartographer compensates for the explorer's limitations.

\subsubsection{Results by Difficulty and Statistical Analysis.}

\begin{table}[htbp]
\centering
\caption{Success Rate (\%) by Difficulty and Statistical Significance}
\label{tab:maze_combined}
\resizebox{\columnwidth}{!}{%
\begin{tabular}{lccccccc}
\toprule
Config & 0\% & 30\% & 60\% & 100\% & $\Delta$Succ. & Wilcoxon $p$ & Cohen's $d$ \\
\midrule
Std EA & 73.0 & 60.0 & 71.7 & 43.3 & \multirow{2}{*}{+3.8\%} & \multirow{2}{*}{0.0119*} & \multirow{2}{*}{0.052} \\
ACE-EA & 79.7 & 65.0 & 71.0 & 47.3 & & & \\
\midrule
Std PSO & 68.3 & 54.7 & 58.3 & 39.0 & \multirow{2}{*}{+27.5\%} & \multirow{2}{*}{$<$0.0001***} & \multirow{2}{*}{0.279} \\
ACE-PSO & \textbf{96.3} & \textbf{95.7} & 77.0 & 61.3 & & & \\
\bottomrule
\multicolumn{8}{l}{\footnotesize * $p<0.05$, *** $p<0.001$. Effect sizes appear modest due to fitness ceiling;} \\
\multicolumn{8}{l}{\footnotesize success rate improvements better indicate practical significance.} \\
\end{tabular}%
}
\end{table}

Table~\ref{tab:maze_combined} stratifies results by difficulty. ACE-PSO achieves 96.3\% and 95.7\% success on low-connectivity mazes---improvements of 28.0 and 41.0 percentage points respectively. Wilcoxon signed-rank tests confirm significance: $p < 0.0001$ for PSO, $p = 0.0119$ for EA, with entirely positive confidence intervals.

\subsubsection{Learned Knowledge.}
Trained GCAs maintain 175--392 non-zero weights among 806 possible transitions. ACE-EA abstracted 11.3 macro-operations per run (10.3 surviving pruning, mean effectiveness 0.254); ACE-PSO showed similar rates (12.0, effectiveness 0.300). These encode 2--5 step reusable path segments. Abstraction rates increase with difficulty: SCMP4 generated 14.6--18.0 macro-operations on average, reflecting longer runs providing more exploration experience.

\section{Conclusion}

This paper proposes ACE, a framework that enhances metaheuristics by coupling them with a learned generative guidance model through three mechanisms: Hebbian weight consolidation learns operation associations via gradient-free updates; guided sampling biases search toward learned patterns; symbolic abstraction extracts persistent patterns into reusable macro-operations.

Experiments demonstrate consistent improvements across both metaheuristics and domains. In molecular design, ACE-EA improves fitness by 10.1\% while solidifying 126 chemically interpretable macro-operations whose transferability is validated through warm-start experiments. In maze navigation, ACE-PSO improves success rate by 27.5\% ($p < 0.0001$) with 49.6\% faster convergence. The explorer-cartographer architecture provides a general template for equipping population-based optimization with cumulative learning.

Limitations include computational overhead and the difficulty of isolating component contributions due to coupling between guidance and feedback. Future work will address: (1) ablation studies via architectural decoupling to quantify marginal impacts of each mechanism; (2) GPU acceleration for efficiency; (3) evaluation on standard numerical benchmarks and constrained optimization; (4) systematic threshold sensitivity analysis; and (5) cross-domain knowledge transfer experiments.

\begin{acks}
This research was partially supported by the Japan Society for the Promotion of Science (JSPS) KAKENHI under Grant 25K00139, and the Tokushima University Tenure-Track Faculty Development Support System, Tokushima University, Japan.
\end{acks}

\bibliographystyle{ACM-Reference-Format}
\bibliography{references}

\end{document}